\documentclass{article}
\usepackage{amsmath,graphicx,mlspconf}
\usepackage{algorithm}
\usepackage{algorithmic} 
\usepackage{graphicx}
\newcommand{\agent}{\textit{agent}}
\newcommand{\environment}{\textit{environment}}
\newcommand{\policy}{\textit{policy}}

\newcommand{\etal}{\textit{et al.}}

\newcommand{\Osa}{\mathcal{O}_{s,a}}
\newcommand{\Csa}{\mathcal{C}_{s,a}}

\title{Online Feature Selection for Activity Recognition using Reinforcement Learning with Multiple Feedback}
\name{Taku Yamagata, Ra{\'u}l Santos-Rodr{\'i}guez, Ryan McConville, Atis Elsts}

\address{
	Department of Engineering Mathematics\\
	University of Bristol\\
	Bristol, UK}

\begin{document}
%\ninept
%

\maketitle
\begin{abstract}
Recent advances in both machine learning and Internet-of-Things have attracted attention to automatic Activity Recognition, where users wear a device with sensors and their outputs are mapped to a predefined set of activities. However, few studies have considered the balance between wearable power consumption and activity recognition accuracy. This is particularly important when part of the computational load happens on the wearable device. In this paper, we present a new methodology to perform feature selection on the device based on Reinforcement Learning (RL) to find the optimum balance between power consumption and accuracy. To accelerate the learning speed, we extend the RL algorithm to address multiple sources of feedback, and use them to tailor the policy in conjunction with estimating the feedback accuracy. We evaluated our system on the SPHERE challenge dataset~\cite{Twomey2016}, a publicly available research dataset. The results show that our proposed method achieves a good trade-off between wearable power consumption and activity recognition accuracy.
\end{abstract}
%
%%%%%%%%%%%%%%%%%%%%%%%%%%%%%%%%%%%%%%%%%%%%%
\section{Introduction}
%%%%%%%%%%%%%%%%%%%%%%%%%%%%%%%%%%%%%%%%%%%%%

% Problem settings...

% Existing approaches...

% Our approach...

% Paper structure...

There is significant interest in understanding activities of daily living (ADL) of people from a wide cross-section of society, but particularly within the healthcare domain. 
This is evidenced by the vast amount of studies undertaken utilizing accelerometers~\cite{doherty2017large,McConville2018c}, perhaps the most commonly used device for detecting ADL. 
%However, for these small battery powered wearable devices, energy efficiency is of the utmost importance and battery life on a single charge is measured in terms of weeks or months~\cite{10.4108/eai.7-9-2017.153063}. Increased efficiency reduces the frequency of device charging, thus decreasing the burden on the wearers, and increasing the amount of data collected.

In the smart home context~\cite{Twomey2016}, low powered wearables constantly transmit their raw data to more computationally powerful devices where the actual processing is carried out. 
However, recently these wearables contain increasingly powerful microcontrollers that are capable of carrying out significant computation. Further, the energy trade-off between on-device computation and transmission is increasingly favouring on-device computation. In fact, when these wearables can adapt to context and make decisions online, energy savings can be made by, for example, dynamically reducing the sample rate of energy expensive wearable heart rate sensors~\cite{McConville2018d}. In a similar vein, Elsts \etal~\cite{Elsts2018} proposed moving a significant step in the activity recognition pipeline to the wearable, demonstrating the significant energy savings that can be made with on-board feature extraction. 

However, one limitation of this work is that the set of features to be extracted are determined a priori.
Intuitively, one can expect that the best set of features, in terms of the energy/accuracy trade-off, to be extracted to accurately recognize a specific activity depends on the context. 
%Consider context \emph{A}, where the wearer has been sleeping for a number of hours, and context \emph{B}, where the wearer is currently climbing the stairs of their home. In context \emph{A} we may find that a low energy set of features are able to reliably detect this activity over long periods of time. In contrast, we may find that in order to correctly detect the correct activity in context \emph{B}, we need a set of features which consumes more energy, for a short duration.

In this work, we propose a Reinforcement Learning (RL) based feature selection approach with the agent running on the wearable. To accelerate the RL algorithm learning, we also introduced a feedback mechanism from the smart home host processor, which has much more processing power and access to extra sensors. The feedback is incorporated into the RL algorithm based on the \textbf{Advise} algorithm ~\cite{Griffith2013} with two extensions - supporting multiple feedback sources and estimating their reliabilities of feedback. These extensions are useful, as the host processor could generate multiple feedback based on the extra sensors, and it is not clear the reliability of each of the feedback source. 
Our main contributions in this work are as follows.
\begin{itemize}
    \item We present a methodology for energy efficient online selection of features from wearables based on context.
    \item We propose a novel RL learning algorithm that extends the work on ~\cite{Griffith2013} by supporting multiple feedback sources and estimating the reliability of feedback in an online fashion.
\end{itemize}

%%%%%%%%%%%%%%%%%%%%%%%%%%%%%%%%%%%%%%%%%%%%%
\section{Related Work}
%%%%%%%%%%%%%%%%%%%%%%%%%%%%%%%%%%%%%%%%%%%%%

There has been much work in the general area of activity recognition~\cite{6365160}, particularly wrist-based accelerometers, due to their performance and acceptability to users~\cite{Towmey2018}. 
However, with the growing popularity of smart homes, potential arose for these wearables to integrate with other devices within the home to benefit from increased context.
%This integration facilitates information fusion~\cite{} to improve the performance and capabilities of activity recognition~\cite{}.
In these settings, features are typically extracted from the raw acceleration data, after data collection, on much more powerful computers than the wearable that collected the data.

However, the transmission energy cost of the raw data is is expensive, and recent work~\cite{Elsts2018} demonstrated that major energy savings can be made by moving the feature extraction to the wearable device, and transmitting features rather than raw data. 
% They measured the power associated with calculating common activity recognition features on these wearables, as well as the performance of activity recognition classifiers trained on these features. 
However, in that work, features are used individually, and are chosen a priori. Our proposed method operates on groups of features, which can change dynamically based on context from other sensors.

The idea of utilizing RL to select features based on cost has been studied previously.
Janisch \etal~\cite{janisch2019classification} pose the task of classification where each feature can be acquired for a cost, and the goal is to optimize the trade-off between the features' costs and classification performance. Similarly, Possas \etal~\cite{Possas2018} utilizes Deep RL to learn a policy to select between two activity recognition methods; one is the motion predictor, using a Long Short-Term Memory (LSTM) network, and the other is a vision predictor using both a Convolutional Neural Network and LSTM. While conceptually similar to our proposed approach, due to their deep nature, they both use considerably more power, and thus unsuitable for the setting of very low-power wearables. Our approach uses simplified discrete states to replace the deep network and employs feedback from other sensors to accelerate the agent's learning speed. 

%The idea of utilizing RL to select features based on cost has been studied previously~\cite{janisch2019classification}. 
%Janisch \etal pose the task of classification where each feature can be acquired for a cost, and the goal is to optimize the trade-off between the features' costs and classification performance. They propose a Deep Reinforcement Learning based approach to solve this resource allocation problem.
%Although it can be applied to the many costs involved in feature selection, it requires multiple predictions of a deep network for each classification. Unfortunately, this approach consumes an impractical amount of power for embedded wearable approaches. 
 
%Similarly, recent work by~\cite{Possas2018} utilizes RL to learn a policy to select between two activity recognition methods; one is the motion predictor, using a Long Short-Term Memory (LSTM) network, and the other is a vision predictor using both a Convolutional Neural Network and LSTM. While conceptually similar to our proposed approach, due to the deep implementations they both use considerably more power, and thus unsuitable for the setting of very low-power wearables. Our approach uses simplified discrete states to replace the deep network and employs feedback from other sensors to accelerate agent's learning speed. 

%%%%%%%%%%%%%%%%%%%%%%%%%%%%%%%%%%%%%%%%%%%%%
\section{Learning Algorithms}
\begin{figure} 
\begin{center}
\includegraphics[width=\columnwidth]{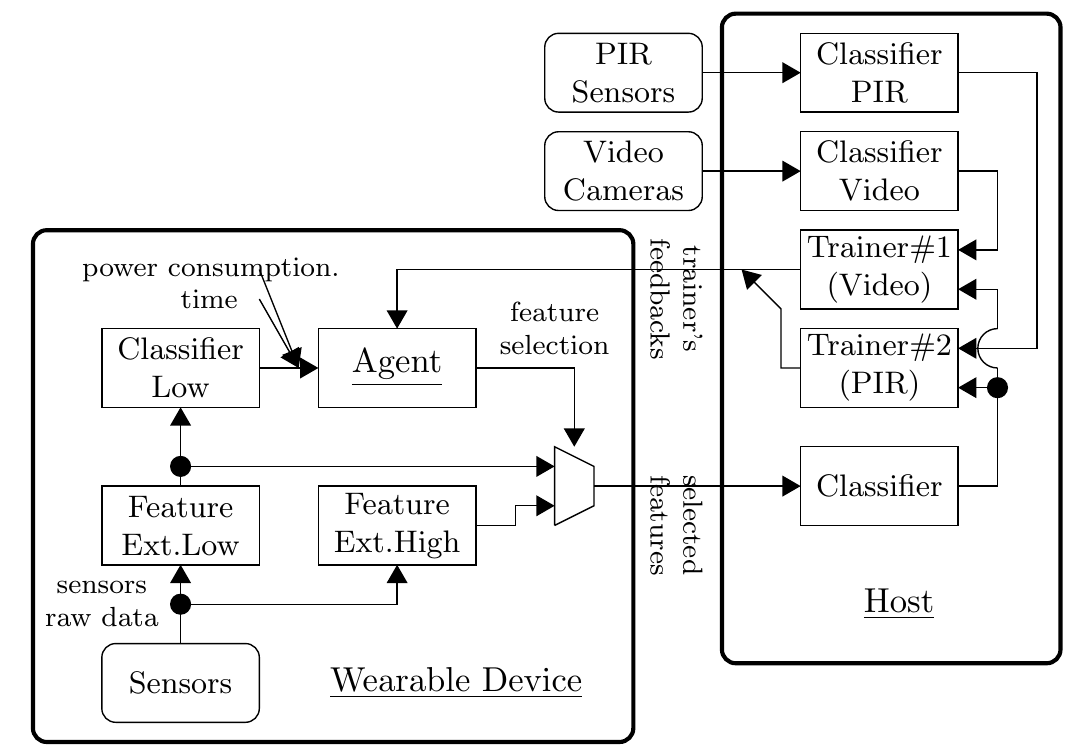}
\caption{Overall structure of the AR system.}
\label{fig:overview}
\end{center}
\end{figure}

%%%%%%%%%%%%%%%%%%%%%%%%%%%%%%%%%%%%%%%%%%%%%
The goal of the our proposed algorithm is to learn a feature selection policy which achieves low power consumption with a low error rate.
We contextualize the method within a smart home setting similar to the SPHERE platform~\cite{Twomey2016}, which consists of numerous different sensor modalities, including PIR and RGB-D cameras, as well as a wrist-worn accelerometer. Fig.~\ref{fig:overview} shows the overall structure of our AR system.
Our wearable extracts features from sensor outputs and transmits the features to the host, which produces predictions of activities. 
To simplify the setting, the wearable has access to two groups of features, one group of low-power features and one group of complex, higher-power features. These are chosen based on their power consumption~\cite{Elsts2018}.
From these, the agent chooses the appropriate set of features to be computed before transmitting them to the host. 
Thus, the aim is to learn the best feature selection policy which achieves a good trade-off between power consumption and accuracy. 

Our method compensates for the lack of computational resources on the device by taking into account feedback from other sensors for each agents action. 
This feedback is then used to shape the agents policy.
Due to the role of each sensor, we will refer to them as \textit{trainers}.
This procedure aids finding the optimum policy and accelerates the agents learning. Furthermore, we extend the existing literature to incorporate multiple trainers and handle unreliable feedback.
In our proposed method and evaluation, the feedback consists of passive infrared sensors (PIR) and RGB-D sensors that are placed throughout a residential home. 
The host processes this data, producing feedback that is transmitted to the wearable agent.

%%%%%%%%%%%%%%%%
\subsection{RL with feedback}
%%%%%%%%%%%%%%%%
The reinforcement learning framework has two components, an \agent{} and an \environment{}~\cite{Sutton1998}. The \agent{}  decides which action to take, and the \environment{} reacts to the action and presents a new state to the \agent{}. The \environment{} also generates rewards, which are special numerical values whose sum the \agent{} tries to maximize over time.

The interactions between the \agent{} and the \environment{} happen in discrete time steps, $t=0,1,2,3...$, and at each time the \agent{} observes the state of the \environment{} ($s_t$) and the reward ($r_t$) and then decides which action ($a_t$) to take next. The goal of the reinforcement learning algorithm is to learn a mapping from states to actions that maximizes the rewards over time through these interactions with the \environment{}. The mapping is called the agent's \policy{}, denoted by $\pi(s,a)$, which indicates the probability of choosing action $a$ in state $s$.

%%%%%%%%%%%%%%%%
\subsubsection{Advise}
%%%%%%%%%%%%%%%%
In order to incorporate feedback into the RL algorithm we build upon the \textbf{Advise} algorithm~\cite{Griffith2013} and thus provide a brief introduction to the approach. 
\textbf{Advise} assumes a binary feedback from a trainer that returns either `right' or `wrong' for a particular agent's choice of action. 
The feedback is accumulated for each state-action pair separately, and it is used to derive a trainer's policy, denoted by $\pi_{F}\left(s,a\right)$, which is then used to modify agent's policy. 
Additionally, $C$ is defined as the probability that the trainer gives the right (consistent) feedback, and assuming a binominal distribution, the trainer's policy is as follows.
\begin{equation} \label{eq220}
\pi_{F}\left(s,a\right) = \frac{ C^{\Delta(s,a)} }{ C^{\Delta(s,a)} + \left( 1-C \right)^{\Delta(s,a)} }
\end{equation}
where $\Delta(s,a)$ is the difference between the number of positive and negative feedback from the trainer. 
%\begin{eqnarray}
%&\Delta(s,a) = h_{s,a}^{+} - h_{s,a}^{-}. \nonumber \\
%&h_{s,a}^{+} : \mbox{number of positive feedback from the trainer}. \nonumber \\
%&h_{s,a}^{-} : \mbox{number of negative feedback from the trainer}. %\nonumber
%\end{eqnarray}
The policy of the trainer is combined with $\pi_{R}\left(s,a\right)$ (policy from the underlying RL  algorithm) by multiplying them together, so that the final policy becomes as
\begin{equation} \label{eq222}
\pi \left(s,a\right) \propto \pi_{F}\left(s,a\right) \times \pi_{R}\left(s,a\right).
\end{equation}
 
%%%%%%%%%%%%%%%%
\subsection{Extensions to Advise}
%%%%%%%%%%%%%%%%
While the \textbf{Advise} algorithm provides a mechanism for incorporating feedback, it has two major limitations. The first is that it requires the consistency level ($C$) prior to receiving feedback, which may be unknown or difficult to estimate in many applications.  
The second limitation is the restriction to a single trainer. 
When multiple trainers are available it may be beneficial to incorporate feedback from all sources.
Further, by estimating their reliability, reliable trainer feedback may be incorporated while avoiding adversarial effects from unreliable trainers.
Thus, in this work we extend \textbf{Advise} to take multiple trainers feedback into account while also treating the consistency level as an unknown parameter, estimating it in an online fashion. 

%%%%%%%%%%%%%%%%
\subsubsection{Consistency Level Estimation}
%%%%%%%%%%%%%%%%
In this section we describe how to estimate the $n_{th}$ trainer's consistency level ($C_{[n]}$). The estimation has two steps, the first step produces an estimate of the consistency level for a given state and action pair for the $n_{th}$ trainer ($\mathcal{C}_{[n] s,a}$), and the second takes an average to obtain an universal $C_{[n]}$ for all state action pairs. In this subsection, all discussions are regarding a single $n_{th}$ trainer, hence we omit the trainer's index $[n]$ from all the variables for simplicity. We will reintroduce the index in the next subsection.

First, we consider estimating the trainer's consistency level for a given state action pair ($\Csa$). With a given number of positive feedback ($h_{s,a}^{+}$) and negative feedback ($h_{s,a}^{-}$) on a given state action pair, this can be derived by maximizing the following log-likelihood function,
\begin{equation} \label{eq301}
l(\Csa)=\log⁡\left( p\left( h_{s,a}^{+}, h_{s,a}^{-};\Csa\right)\right).
\end{equation}
As there is no model to compute this likelihood function, we introduce a hidden parameter $\Osa$, and then marginalize out the hidden parameter to obtain the original likelihood function. The hidden parameter is a boolean that is 1 when $a$ is the optimal action at state $s$, and 0 when it is not. Eq.~\ref{eq301} can be rewritten as:  
\begin{equation} \label{eq302}
l(\Csa)=\log⁡\left( \sum_{\Osa} p\left( h_{s,a}^{+}, h_{s,a}^{-}, \Osa;\Csa\right)\right).
\end{equation}
% EM ALGORITHM (M-STEP)
We then use the Expectation Maximization (EM) algorithm~\cite{Dempster1977} to compute a maximum likelihood estimate of the consistency level ($\Csa$). The $i_{th}$ iteration of the M-step can be written as follows,
\begin{equation} \label{eq304}
\Csa^{\left(i+1\right)} = \frac{P_{1} \cdot h_{s,a}^{+} + P_{0} \cdot h_{s,a}^{-}}{h_{s,a}^{+} + h_{s,a}^{-}}
\end{equation}
where $\Csa^{\left(i+1\right)}$ is the estimated consistency level at the $i_{th}$ iteration, $P_0$ and $P_1$ are given as follows:
\begin{equation} \label{eq305}
\begin{split}
P_0 = p\left( \Osa=0 | h_{s,a}^{+}, h_{s,a}^{-};\Csa^{\left(i\right)}\right). \\
P_1 = p\left( \Osa=1 | h_{s,a}^{+}, h_{s,a}^{-};\Csa^{\left(i\right)}\right).
\end{split}
\end{equation}
The E-step fundamentally requires computing Eq.~\ref{eq305}, using Eq.~\ref{eq220} and the probabilities derived from interaction with the environment. $P_{1}^{Q}(s,a)$ and $P_{0}^{Q}(s,a)$ are the probabilities of the optimal and non optimal action. As a result, they can be written as follows.

\begin{equation} \label{eq307}
\begin{split}
P_0 = \frac{ P_{0}^{Q}(s,a)  \cdot \left(1-\Csa^{(i)}\right)^{\Delta(s,a)} }{ P_{1}^{Q}(s,a) \cdot \left(\Csa^{(i)}\right)^{\Delta(s,a)} + P_{0}^{Q}(s,a) \cdot \left(1-\Csa^{(i)}\right)^{\Delta(s,a)} }. \\
P_1 = \frac{ P_{1}^{Q}(s,a) \cdot \left(\Csa^{(i)}\right)^{\Delta(s,a)} }{ P_{1}^{Q}(s,a)  \cdot \left(\Csa^{(i)}\right)^{\Delta(s,a)} + P_{0}^{Q}(s,a)  \cdot \left(1-\Csa^{(i)}\right)^{\Delta(s,a)} }.
\end{split}
\end{equation}
We set $P_{1}^{Q}(s,a)=\pi_R(s,a)$ and $P_{0}^{Q}(s,a)=1-\pi_R(s,a)$. The algorithm is summarized in Algorithm~\ref{alg1}.

% ALGORITHM for Consistency Level Estimation
\begin{algorithm}
\caption{Consistency Level Estimation}
\label{alg1}
\algsetup{
linenosize=\small,
linenodelimiter=.
}
\begin{algorithmic}[1]
\REQUIRE $P_{0}^{Q}(s,a)$, $P_{1}^{Q}(s,a)$, $h_{s,a}^{+}$ and  $h_{s,a}^{-}$ 
\STATE $\Delta(s,a) \leftarrow h_{s,a}^{+} - h_{s,a}^{-}$
\STATE $i \leftarrow 1$
\STATE $C^{(i)} \leftarrow 0.5$
\WHILE{TRUE}
  \STATE $P_0 \leftarrow \frac{ P_{0}^{Q}(s,a)  \cdot \left(1-C^{(i)}\right)^{\Delta(s,a)} }{ P_{1}^{Q}(s,a) \cdot \left(C^{(i)}\right)^{\Delta(s,a)} + P_{0}^{Q}(s,a) \cdot \left(1-C^{(i)}\right)^{\Delta(s,a)} }$
  \STATE $P_1 \leftarrow \frac{ P_{1}^{Q}(s,a) \cdot \left(C^{(i)}\right)^{\Delta(s,a)} }{ P_{1}^{Q}(s,a)  \cdot \left(C^{(i)}\right)^{\Delta(s,a)} + P_{0}^{Q}(s,a)  \cdot \left(1-C^{(i)}\right)^{\Delta(s,a)} }$
  \STATE $C^{\left(i+1\right)} \leftarrow \frac{P_{1} \cdot h_{s,a}^{+} + P_{0} \cdot h_{s,a}^{-}}{h_{s,a}^{+} + h_{s,a}^{-}}$
  \IF{$C^{(i+1)} == C^{(i)}$}
    \STATE break
  \ENDIF
  \STATE $i \leftarrow i+1$
\ENDWHILE
\RETURN $C^{(i)}$
\end{algorithmic}
\end{algorithm}
% FIXED LEARNING RATE
Now that we have derived an algorithm to estimate consistency level for each state-action by using the EM algorithm, we are going to summarize it over the state-action space to come up with one consistency level value for each trainer.
%Although we assume that the trainer consistency level is independent of the state and action pairs, it is not appropriate to just average $\Csa$ over all possible state and actions, because there are many state-action pairs rarely visited and the estimate of $\Csa$ for these pairs is not accurate. It is sensible to take a weighted average over the state-action pairs that actually are being experienced with the current policy. %as it is more likely to have more accurate the consistency level estimations, and also these consistency levels actually affect the result of interacting with an environment.
In order to compute the consistency level, we run the recursive averaging method shown in Eq~\ref{eq310} for every state-action pair actually experienced by the agent.
\begin{equation} \label{eq310}
C = C + \alpha \cdot ( \Csa - C )
\end{equation}
where $\alpha \in [0,1]$ is the learning rate, $C$ is averaged consistency level and $\Csa$ is the estimated consistency level for the current state-action pair.

% ADAPTIVE LEARNING RATE
Additionally, we consider an approach which adaptively changes the learning rate based on the ratio of accuracy of the estimated $\Csa$ and the averaged $C$ as follows,
\begin{equation} \label{311}
\alpha = \alpha_{0} \cdot \frac{ \mbox{ Accuracy of } \Csa }{ \mbox{Accuracy of } C}
\end{equation}
where $\alpha_{0}$ is a \textbf{base learning rate}, which is fixed and scaled by the ratio of the accuracies.
The consistency level estimation uses two sources, information from the underlying RL algorithm ($P_{1}^{Q}$ and $P_{0}^{Q}$) and the trainer's feedback ($h_{s,a}^{+}$ and $h_{s,a}^{-}$). Thus, we estimate these accuracies separately and combine them by multiplying them together to get the accuracy of the consistency level estimation.
For the underlying reinforcement learning accuracy, we use as a metric the absolute value of state-action value function (Q function), added up over all actions.
\begin{equation} \label{312}
\mathcal{Q}(s) = \sum_{a \in A} |Q(s,a)|.
\end{equation}
For the trainer's feedback accuracy, we simply use the amount of feedback or the given state, i.e.,
\begin{equation} \label{313}
\mathcal{H}(s) = \sum_{a \in A} h_{s,a}^{+} + h_{s,a}^{-}.
\end{equation}
The above metrics are used for estimating the accuracy of $\Csa$, and we use the following recursive averaging update to track these metrics for the averaged consistency level $C$.
\begin{equation}
\begin{split} \label{314}
\tilde{\mathcal{Q}} = \tilde{\mathcal{Q}} + \alpha (  \mathcal{Q}(s) - \tilde{\mathcal{Q}} ). \\
\tilde{\mathcal{H}} = \tilde{\mathcal{H}} + \alpha (  \mathcal{H}(s) - \tilde{\mathcal{H}} ).
\end{split}
\end{equation}
Then, we calculate the learning rate $\alpha$ by using $\mathcal{Q}(s), \mathcal{H}(s), \tilde{\mathcal{Q}}$ and $\tilde{\mathcal{H}}$.
\begin{equation} \label{4f}
\alpha = \alpha_{0} \cdot \frac{ \mathcal{Q}(s) \cdot \mathcal{H}(s) }{ \tilde{\mathcal{Q}} \cdot \tilde{\mathcal{H}} }.
\end{equation}
As $\alpha \in [0,1]$, we limit the upper value of $\alpha$ to be 1.0 by simply taking $\min(\alpha, 1.0)$. 
For our evaluation we fix the base learning rate ($\alpha_{0}$) to $1.0/16.0$ as in practice it represents a good trade-off between the overall learning speed and suppressing noise in the consistency level estimation.

% ALGORITHM for Adaptive Learning Rate
\begin{algorithm}
\caption{Consistency Level Estimation with an Adaptive Learning Rate}
\label{alg2}
\begin{algorithmic}[1]
\REQUIRE $\alpha_{0}$, $Q(s,a)$, $h_{s,a}^{+}$ and $h_{s,a}^{-}$
\REQUIRE $C$, $\tilde{\mathcal{Q}}$ and $\tilde{\mathcal{H}}$ persistent variables ($C$ initialized 0.5, $\tilde{\mathcal{Q}}$ and %$\tilde{\mathcal{H}}$ initialized 0.1 or small positive number)
\STATE $\mathcal{Q}(s) \leftarrow \sum_{a' \in A} |Q(s,a')|$
\STATE $\mathcal{H}(s) \leftarrow \sum_{a' \in A} h_{s,a'}^{+} + h_{s,a'}^{-}$
\STATE $\alpha \leftarrow \frac{ \mathcal{Q}(s) \cdot \mathcal{H}(s) }{ \tilde{\mathcal{Q}} \cdot \tilde{\mathcal{H}} } \cdot \alpha_{0}$
\STATE $C \leftarrow C + \alpha \cdot ( C(s,a) - C )$
\STATE $\tilde{\mathcal{Q}} = \tilde{\mathcal{Q}} + \alpha (  \mathcal{Q}(s) - \tilde{\mathcal{Q}} ) $
\STATE $\tilde{\mathcal{H}} = \tilde{\mathcal{H}} + \alpha (  \mathcal{H}(s) - \tilde{\mathcal{H}} ) $
\RETURN $C$
\end{algorithmic}
\end{algorithm}

%%%%%%%%%%%%%%%%
\subsubsection{Multiple Trainers}
%%%%%%%%%%%%%%%%
 In order to incorporate multiple trainers, we assume each trainer has a different consistency level, and the $n_{th}$ trainer's consistency level is denoted by $C_{[n]}$. The Bayes optimal method to combine probabilities from (conditionally) independent sources is multiplying them together~\cite{Bailer-Jones2011}, hence the policy for overall multiple trainers $\pi_{F}(s,a)$ can be derived as follows by  employing each trainer's policy given in Eq.\ref{eq220},
\begin{equation}
\label{eq221}
\pi_{F}(s,a) \propto \prod_{n=1}^{N} \left(C_{[n]}\right)^{\Delta_{[n]}(s,a)}
\end{equation}
where $N$ is the number of trainers, and $\Delta_{[n]}(s,a)$ is the difference between positive and negative feedback on the state $s$ and action $a$ from the $n_{th}$ trainer. 
%The algorithm is summarized in Algorithm ~\ref{alg2}.
%For the consistency level estimation, we just need to learn it separately for each trainer. To do so, we need to store the number of positive feedbacks and negative feedbacks separately for each trainer, and use the same method as described above.

%%%%%%%%%%%%%%%%%%%%%%%%%%%%%%%%%%%%%%%%%%%%%
\section{Evaluation}
%%%%%%%%%%%%%%%%%%%%%%%%%%%%%%%%%%%%%%%%%%%%%
%\subsection{Dataset}
To evaluate our proposed approach we use the SPHERE challenge dataset~\cite{Twomey2016}.
This publicly available dataset consists of human annotated activity labels from a variety of different sensors, aligning with our proposed setting.
Importantly for this work, it contains acceleration data from a wrist-worn wearable sampled at 20Hz ($\pm$ 4G) from ten different participants, each following the same script within a residential house. 
We use the complete set of annotated 20 activities, which includes ambulation activities (e.g., walking, jumping), posture activities (e.g., standing, sitting) and transition activities (e.g., sit to stand, turning).
The dataset also contains data from RGB-D cameras, which were placed in multiple rooms of the home, as well as passive infrared (PIR) sensors, both of which will be used as feedback sources. 
We reprocess the data to assign activity labels at the one second granularity.

\subsection{RL Setup}
RL defines a class of algorithms for solving a Markov Decision Process (MDP). A MDP is defined by the tuple $\left(S,A,T,R,\gamma\right)$ for the set of possible states $S=\left(t, \mathcal{P}, c_{low}\right)$, where $t$ is the elapsed time in minutes, $\mathcal{P}$ is power consumption in mC and $c_{low}$ is the low energy feature classifier output. Because $t$ and $\mathcal{P}$ are rounded to the closest integer, our state space is discrete. $A$ is the set of actions, either using low energy features or high energy features. The low energy feature set includes four time domain features, namely the mean, min and max, as well as the number of zero crossings,  while the high energy features are the low energy features plus higher energy features such as quartiles, histograms and spectral features. $T$ is the state transition function. Finally, $R$ is a reward function that is all zeroes except for the last state in the episode. The final state reward is calculated based on the total power consumption and average error rate in the episode as Eq.~\ref{eqReward}.
\begin{equation} \label{eqReward}
r = - \lambda \cdot p_e - (\mathcal{P} / \mathcal{P}_{tgt})^2 
\end{equation}
where $p_e$ is the error rate, which is the percentage of activities misclassified ($1 - accuracy$), $\mathcal{P}$ is the power consumption, $\mathcal{P}_{tgt}$ is the desired power consumption and $\lambda$ is a positive real number controlling trade off between error rate and power consumption. In our experiment we set  $\mathcal{P}_{tgt}=16.7$mC and $\lambda=1.0$. $16.7$mC is derived by assuming 10mAh battery charge is allocated for feature extraction over a 30 day period. The reward function quadratically penalizes the normalized power consumption. We define the problem as an episodic task where each task is 20 minutes long, with a time step of 5 seconds. For the sake of simplicity, the Agents use \textit{Q-Learning} (QL)~\cite{Watkins1989} and \textit{Boltzmann exploration policy}. The hyper parameters are the discount factor $\gamma=0.99$, the learning rate $\alpha=0.1$ and the temperature parameter for the Boltzmann exploration $\tau=0.1$.

The two trainers are implemented on the host to generate feedback for the RL agent on the wearable. Each trainer has a classifier from one of the additional sensors, namely PIR or RGB-D cameras. The feedback is generated by comparing the extra sensor classification and the selected feature classification. If the low energy feature set classification result is same as the extra sensor classification result, positive feedback is given for low energy feature set and negative feedback for high energy feature set. If the low energy feature classification result is not same as the extra sensor classification result, we check if the high feature classification result is the same as the extra sensor classification result. If they match, and the current power consumption is still less than the target power consumption, it generates positive feedback for high feature set. Otherwise it does not generate any feedback.
%\todo{Can you summarize algorithm below in sentence or two? - for space?}
% The detailed feedback generation algorithm is described in Algorithm~\ref{alg3}.
% \begin{algorithm}
% \caption{Feedback Generation}
% \label{alg3}
% \begin{algorithmic}[1]
% \REQUIRE -\\
% $action$ agent's action (`low' or `high')\\
% $\mathcal{P}$ power consumption in the episode\\
% $\mathcal{P}_{tgt}$ power consumption target\\
% $c_{ext}$ extra sensor (PIR or RGB-D) classifier result\\
% $c_{low}$, $c_{high}$ low and high energy classifier result\\
% \IF{$c_{low}$ == $c_{ext}$}
%     \IF{action == `low'}
%         \item feedback = `right'
%     \ELSE
%         \item feedback = `wrong'
%     \ENDIF
% \ELSE
%     \IF{$action==$`high' and $c_{high}==c_{ext}$ and \\
%         \ \ \ $\mathcal{P}<\mathcal{P}_{tgt}$}
%         \item feedback = `right'
%     \ELSE
%         \item feedback = none
%     \ENDIF
% \ENDIF
% \RETURN feedback
% \end{algorithmic}
% \end{algorithm}

\subsection{Results}
We will compare our proposed method against three different approaches, including two baselines. The first baseline is the use of high or low power features chosen at random with equal probability (Random). The second baseline will only use low power features (Fixed Low). 
We will also compare our method that uses multiple trainers with online consistency level estimation (Multi-Trainers) with one using Q-Learning (QL), which is commonly used with RL, often as a baseline. 

\begin{figure} 
\centering
\includegraphics[width=0.85\columnwidth]{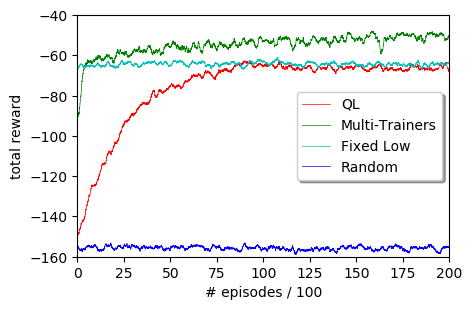}
\caption{Reward Learning Curves.}
\label{fig:plotRW}
\includegraphics[width=0.85\columnwidth]{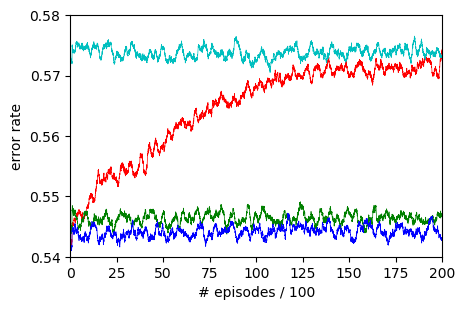}
\caption{Error Rate Learning Curves}
\label{fig:plotEr}
\includegraphics[width=0.8\columnwidth]{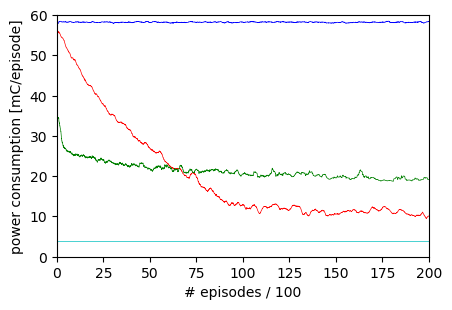}
\caption{Power Consumption Learning Curves}
\label{fig:plotPC}
\includegraphics[width=0.8\columnwidth]{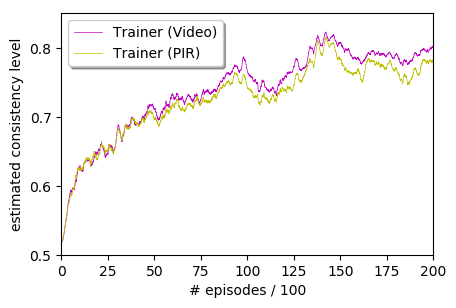}
\caption{Consistency Level Learning Curves}
\label{fig:plotC}
\end{figure}

The learning curves for each different approach can be seen in Figure~\ref{fig:plotRW}, showing the total reward obtained each episode. This figure shows that our proposed method learns quicker and reaches a higher asymptote compared to QL, while the random baseline maintains remains stagnant at around $-155$ and the low power features are around the level that QL reaches after 10,000 episodes. Figure~\ref{fig:plotEr} and Figure ~\ref{fig:plotPC} show how each algorithm learns the trade-off between the power consumption and error rate, which is the percentage of activities misclassified ($1 - accuracy$.) 
 For reference, the performance of a classifier which labelled activities randomly would have an error rate of 0.81. 
 It is clear that the Fixed Low feature approach has consistently the highest error rate, with QL approaching a similar level after around 15,000 episodes. 
The two approaches consistently maintaining a low error rate are the Multi-Trainers and the random approach, with the random approach typically having a slightly lower error rate. However, Figure~\ref{fig:plotPC} shows that the random approach has a significantly higher power consumption than all other methods, including over twice the power consumption of our Multi-Trainers method.
While the Fixed Low consistently has a lower power consumption, it has the highest error rate and thus does not achieve a good balance. 
QL eventually reduces the power consumption to a level lower than our method, however as this occurs the error rate increases. Therefore, in comparison to the others, our method maintains a consistently low power consumption, while maintaining a low error rate.
%We believe the reason for the poorer performance of QL is due to the error rate measurement being noisy, and thus there is little benefit in reducing it. Therefore, the agent tends to learn that reducing its power consumption leads to higher rewards. %On the contrary, intuitively, our approach helps the agent to `notice' that it is rewarded for reducing the error rate by guiding its policy to use high energy features in difficult contexts by introducing feedback from the host. 
Finally, the learning curve for the consistency level is plotted in Figure ~\ref{fig:plotC}. It shows that, as the agent learns the environment, the agent perceives that the feedback is consistent with what it has learned, leading to to an increase in the consistency level. Interestingly, it also shows a slightly higher consistency level for the RGB-D sensor based trainer, which is expected as RGB-D sensors are thought to provide more information than PIR sensors.

%%%%%%%%%%%%%%%%%%%%%%%%%%%%%%%%%%%%%%%%%%%%%
\section{Conclusion}
%%%%%%%%%%%%%%%%%%%%%%%%%%%%%%%%%%%%%%%%%%%%%
In this work we proposed a method for online feature selection for sensor streams by leveraging feedback from multiple trainers (alternative sensors) while estimating their consistency. We evaluated our approach on a publicly available activity recognition dataset, where the task was to reduce the energy consumption of a wearable device containing the RL agent while maintaining a low error rate. 
 The evaluation demonstrated that our proposed method was the only approach able to maintain the error rate while reducing power consumption, thus achieving our objective. The baselines failed to achieve the balance between the error rate and power consumption. The random policy achieved the lowest error rate by consuming the highest power, and the fixed low policy has the lowest power consumption by sacrificing the error rate. Q-Learning learned a policy in-between the previous two baselines, but fails to learn a policy to achieve the higher reward by reducing the error rate with small increments of its power consumption.
%\todo{TODO sentence in here summarising how baselines etc. performed in comparison.)}
We believe that this is because the error rate results have higher variance compared to the averaged error rate difference due to the feature selection, hence it is difficult to see the benefit of reducing the error rate by selecting the high energy feature set. %On the other hand, the power consumption results have low variance, so it is easy to see the benefit of reducing the power consumption. As result, Q-Learning tends to converge towards the fixed low policy. 
%\todo{TODO this sentence is confusing. Can wee add another setence explaining the improvement in performance?}
Further, we found our motivation for learning the consistency level to be justified by the experiments.
The RGB-D sensor, which would be expected to have a high consistency level, empirically was close to the PIR sensor. 
Thus, setting this consistency level a priori would not have been straightforward or optimal.
%Additionally, this suggests that even with access to accurate sensors the correct feedback may not always be generated.
 We found that our method was robust to unreliable feedback and able to provide guidance to the agent to explore in the correct direction, ultimately achieving the best trade-off.
% The consistency level estimation for RGB-D sensor is not particularly high, suggesting that even when the host has access to more accurate sensors it still cannot consistently generate the correct feedback.
%It might be possible to improve our feedback generation algorithm to increase its accuracy, however, it might always have a risk of over-fitting. 
% We note 
% All of above suggest the importance of estimating the consistency level, as it is difficult to know a priori from the context.  

\bibliographystyle{ieeebib.bst}
\bibliography{my_library}

\end{document}